\documentclass[11pt]{article}

\usepackage[final]{acl} 

\usepackage{times}
\usepackage{latexsym}

\usepackage[T1]{fontenc}

\usepackage[utf8]{inputenc}
\usepackage{amsmath}

\usepackage{microtype}

\usepackage{inconsolata}

\usepackage{graphicx}

\usepackage{booktabs}
\usepackage[table]{xcolor}
\usepackage{arydshln}
\usepackage[most]{tcolorbox}
\usepackage{amssymb}
\usepackage{amsthm}

\definecolor{myBlue}{RGB}{233,242,255}   
\definecolor{myOrange}{RGB}{255,242,229} 

\newcounter{mybox}

\usepackage[ruled,vlined,linesnumbered]{algorithm2e}
\usepackage{float}

\usepackage{amsfonts}

\definecolor{myblue}{HTML}{5e9ec8}
\definecolor{mygreen}{HTML}{7fc77e}
\definecolor{myorange}{HTML}{fb933c}
\usepackage{fontawesome5}
\usepackage{marvosym}

\newlength{\colskip}
\usepackage{multirow}

\title{No More Stale Feedback:\\ Co-Evolving Critics for Open-World Agent Learning}

\author{
 \textbf{Zhicong Li\textsuperscript{1,2}\footnotemark[1]},
 \textbf{Lingjie Jiang\textsuperscript{3}\footnotemark[1]},
 \textbf{Yulan Hu\textsuperscript{2}},
 \textbf{Xingchen Zeng\textsuperscript{4}},
\\
 \textbf{Yixia Li\textsuperscript{5}},
 \textbf{Xiangwen Zhang\textsuperscript{2}},
 \textbf{Guanhua Chen\textsuperscript{5}},
\\
 \textbf{Zheng Pan \textsuperscript{2}},
 \textbf{Xin Li\textsuperscript{2}},
 \textbf{Yong Liu\textsuperscript{1}\footnotemark[2]}
\\
\\
 \textsuperscript{1}Gaoling School of Artificial Intelligence, Renmin University of China,\\
 \textsuperscript{2}Amap, Alibaba Group, \textsuperscript{3}Peking University, \\
 \textsuperscript{4}The Hong Kong University of Science and Technology (Guangzhou),\\
 \textsuperscript{5}Southern University of Science and Technology
 \\
\small
\texttt{\{zhicongli, liuyonggsai\}@ruc.edu.cn, lingjiejiang@stu.pku.edu.cn}\\
\small \texttt{\{huyulan, zhangxiangwen.zxw, panzheng.pan, beilai.bl\}@alibaba-inc.com}\\
\small \texttt{xzeng159@connect.hkust-gz.edu.cn, liyixia@me.com, ghchen08@gmail.com}
\\
}

\begin{document}
\maketitle
\footnotetext[1]{Equal contribution.}
\footnotetext[2]{Corresponding author.}
\begin{abstract}

Critique-guided reinforcement learning (RL) has emerged as a powerful paradigm for training LLM agents by augmenting sparse outcome rewards with natural-language feedback. However, current methods often rely on static or offline critic models, which fail to adapt as the policy evolves. In on-policy RL, the agent's error patterns shift over time, causing stationary critics to become stale and providing feedback of diminishing utility. 
To address this, we introduce \textbf{ECHO}{ (\textbf{E}volving \textbf{C}ritic for \textbf{H}indsight-Guided \textbf{O}ptimization)}, a framework that jointly optimizes the policy and critic through a synchronized co-evolutionary loop. ECHO utilizes a cascaded rollout mechanism where the critic generates multiple diagnoses for an initial trajectory, followed by policy refinement to enable group-structured advantage estimation. We address the challenge of learning plateaus via a saturation-aware gain shaping objective, which rewards the critic for inducing incremental improvements in high-performing trajectories. By employing dual-track GRPO updates, ECHO ensures the critic's feedback stays synchronized with the evolving policy. Experimental results show that ECHO yields more stable training and higher long-horizon task success across open-world environments.
\end{abstract}

\section{Introduction}
Reinforcement learning~\citep{sutton1998reinforcement} has emerged as a promising paradigm for training Large Language Model (LLM)-based agents~\citep{anthropic2024claude, team2025kimi}, enabling them to navigate complex tasks through environmental interactions. 
Within this paradigm, reward signals~\citep{wen2025reinforcement} serve as the fundamental compass for policy optimization. However, these signals often lack actionability, as they merely reflect the final outcome without providing the diagnostic insights necessary for effective refinement, ultimately leading to significant data inefficiency~\citep{gao2025llm, yang2025lighthouse}.


To bridge this gap, recent research has introduced \textit{linguistic critics} to provide diagnostic feedback~\citep{dhuliawala2024chain}. A common line of work uses template-based critiques~\citep{wang2025hint, liu_ghpo_2025, huang2025boosting}, which are 
computationally inexpensive but lack the adaptability to deliver feedback tailored to the agent’s specific actions.
To provide more targeted guidance, another line of work employs independently fine-tuned, separate critic models to refine policy outputs \citep{mcaleese2024llm}. 
These models are typically designed to act as external supervisors, aiming to provide the  diagnostic feedback necessary to resolve complex failures.

Although these methods overcome the limitations of static templates by offering more detailed feedback, they remain decoupled from the policy's learning process, \textit{implicitly assuming that the optimal critique strategy is stationary}. In on-policy RL, however, the policy continuously evolves, inducing a shifting trajectory distribution and a corresponding drift in failure patterns: early-stage rollouts may be dominated by coarse mistakes that benefit from high-level hints, whereas later-stage policies are more often bottlenecked by subtle, hard-to-localize defects.
Consequently, a critic trained (and then frozen) on an earlier distribution can become \textit{stale}, producing feedback that is redundant, miscalibrated in granularity, or even misleading for the current policy, and causing its marginal utility to decay as training progresses. This critic staleness fundamentally limits sample efficiency and prevents critique-guided RL from sustaining improvement in long-horizon refinement.



\begin{figure*}[t]
  \centering
  \includegraphics[width=\textwidth]{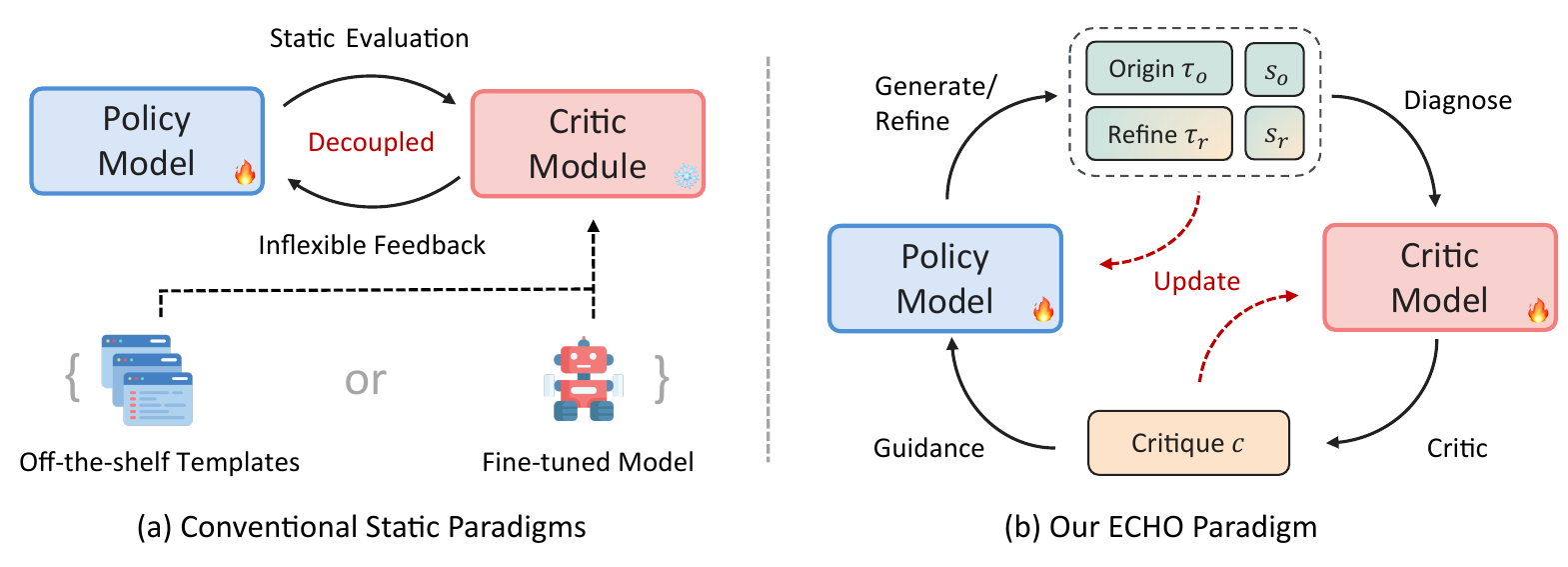}
  \vspace{-7mm}
  \caption{Comparison of critic paradigms. 
  (a) \textbf{Conventional Static Paradigms}: Use decoupled, frozen critic modules initialized from off-the-shelf templates or fine-tuned separate models, resulting in static evaluation and inflexible feedback. 
  (b) \textbf{Our ECHO Paradigm}: Policy and critic co-evolve organically. 
  The policy first generates an initial rollout $\tau_o$, refined to $\tau_r$ using the critic’s diagnostic guidance $c$. 
  Both models are jointly updated, ensuring the critic's diagnostic precision synchronizes with the policy's evolving failure patterns.}
  \label{fig:intro}
\end{figure*}

Motivated by this observation, {we posit that the critic should be treated as a co-evolving module rather than a stationary supervisor, adapting alongside the policy (Figure~\ref{fig:intro})}. Concretely, we propose ECHO (\textbf{E}volving \textbf{C}ritic for \textbf{H}indsight-Guided \textbf{O}ptimization), a framework that fosters a symbiotic optimization loop between the policy and the critic. 
Instead of rewarding the critic for sounding plausible, we directly optimize it for {policy improvement}: critiques are evaluated by the performance gains they induce after refinement, and the critic is updated in lockstep with the policy to track its changing failure modes. To make this co-evolution stable and sample-efficient, ECHO employs a cascaded diagnostic-and-corrective rollout that generates group-structured trajectories for relative advantage estimation, and introduces a saturation-aware gain shaping to provide informative learning signals even when improvements become incremental.


Our main contributions are: (1) We identify and empirically demonstrate critic staleness in critique-guided RL, freezing the critic leads to a clear decay in critique utility as the policy improves. 
(2) We introduce ECHO, a synchronized co-evolutionary optimization paradigm that jointly aligns the critic and the policy via dual-track GRPO. 
(3) We propose a saturation-aware reward design and group-relative optimization scheme that jointly improve training stability and boost performance across tasks.

\section{Related Work}
In long-horizon decision-making for LLM-based agents, scalar outcome rewards are often non-diagnostic, motivating language-based critiques as actionable supervision~\citep{gao2025llm, yang2025lighthouse,zhao2025mas}.
Prior work typically implements language critics either as static, template/offline-generated feedback, or as separately trained critic models.
\paragraph{Template-based Critics.} 
A lightweight line of work injects {pre-defined} hints as critique signals, avoiding training a separate critic model. 
HINT~\citep{wang2025hint} steers ineffective rollouts toward effectiveness by appending generic, hand-crafted hints to trigger regeneration.
\citet{tang2025not} further adopts a small set of error-conditioned prompt templates, routing different failure cases to different pre-defined guidance patterns.
Moving beyond generic guidance, LUFFY~\citep{yan2025learning} mitigates inefficient exploration by injecting a teacher model’s correct answer as the rollout outcome.
To better control the granularity of the guidance, more structured hints have also been explored. 
GHPO~\citep{liu_ghpo_2025} and ADHint~\citep{zhang2025adhint} provides stronger supervision by injecting masked partial reference solutions as hints, effectively revealing part of the answer to stabilize and accelerate learning.

StepHint~\citep{zhang2025stephint} uses a teacher model to generate a full chain-of-thought, splits it into $N$ reasoning steps, and forms hints by concatenating different numbers of prefix steps. 
In contrast, Scaf-GRPO~\citep{zhang2025scaf} designs critic templates that progress from abstract to concrete guidance, providing coarse-to-fine guidance conditioned on the model's current performance.

\paragraph{Training-based Critics.}
Another line of work trains dedicated critic models to generate more informative, diagnostic feedback.
Early attempts~\citep{saunders2022self, ke2024critiquellm, xi2024enhancing, tangself} primarily rely on single-stage fine-tuning, typically by curating critique datasets and training models to generate natural-language feedback for evaluation and verification.
\citet{yu2025training} propose Refinement-oriented Critique Optimization (RCO), which trains a critic in a critique–refinement loop by rewarding critiques according to the utility of the actor’s refined outputs.
Multi-stage training has also been investigated to stabilize learning across different training objectives. 
CGI~\citep{yang2025lighthouse} leverages critique-guided iterative improvement for agents through staged updates, typically treating the critic as a fixed supervisor. 
CTRL~\citep{xie2025teaching} introduces a two-stage training pipeline that first distills critiques via SFT and then applies GRPO to optimize critique generation directly for downstream refinement success.


Despite these advances, most training-based critics are trained off-policy and then frozen or updated asynchronously, remaining decoupled from on-policy policy learning. 
As the policy’s trajectory distribution and failure patterns shift over time, the critic becomes stale, and its ability to provide useful critiques gradually decays.

\section{Methodology}
\begin{figure*}[t]
  \centering
  \includegraphics[width=0.95\textwidth]{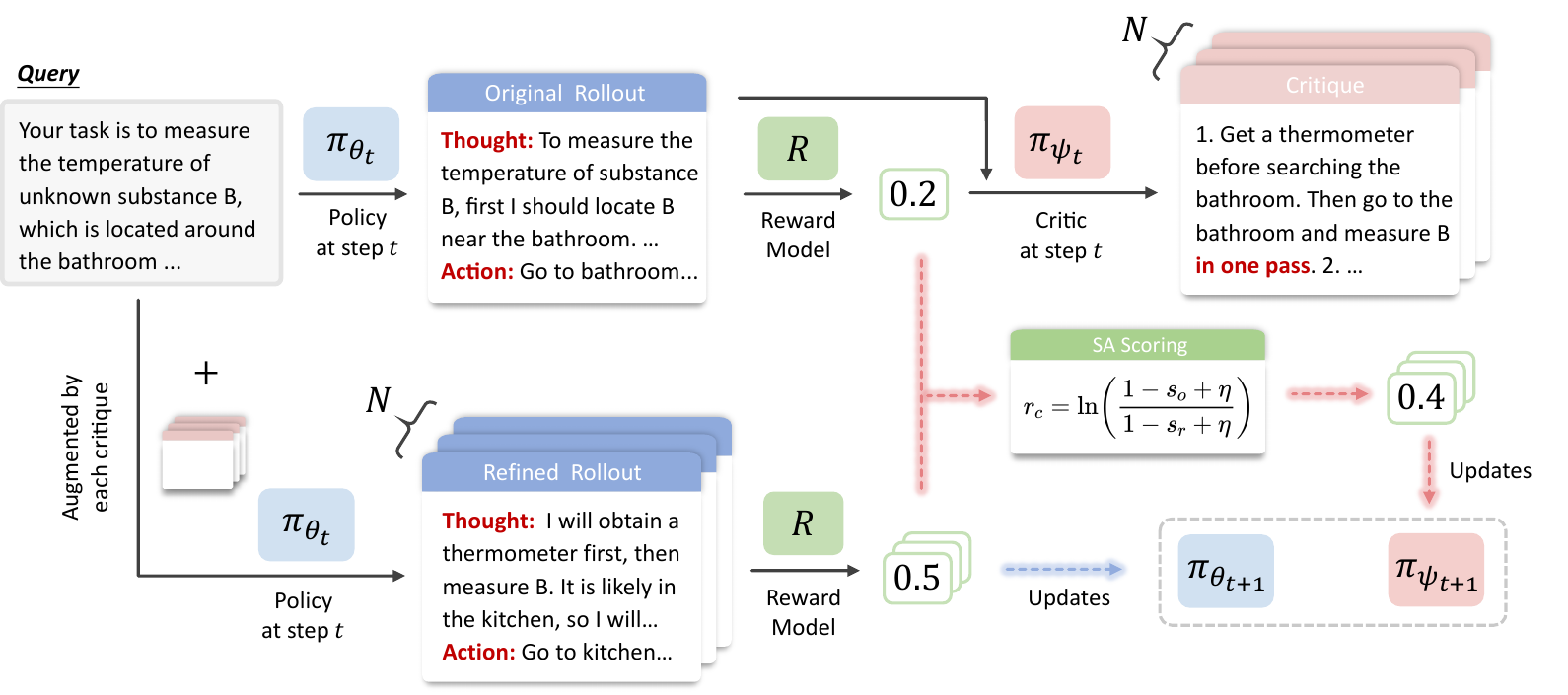}
  \vspace{-1mm}
  \caption{Overview of ECHO training with saturation-aware (SA) critic rewards. At step \(t\), the policy \(\pi_{\theta_t}\) produces rollouts \(\tau_o\), which are scored by a reward model to obtain \(s_o\). A critic \(\pi_{\psi_t}\) generates critiques that are appended to the original query to elicit refined rollouts \(\tau_r\), scored as \(s_r\). We compute the SA critic reward \(r_c\) to emphasize last-mile improvements near saturation, and update the critic and policy synchronously to obtain \(\pi_{\psi_{t+1}}\) and \(\pi_{\theta_{t+1}}\).}
  \label{fig:method}
\end{figure*}

To address critic staleness caused by decoupled training under on-policy failure-pattern drift, we propose ECHO, a co-evolutionary interplay between a Policy $P_\theta$ and a Critic $C_\psi$, rather than a static supervision task. Within this paradigm, we treat the refinement process as a dynamic synchronization problem where two models co-evolve in a shared on-policy trajectory space:
\vspace{-1mm}
\begin{itemize}
    \item $P_\theta$ (The Actor) learns to convert diagnostic feedback into  corrective actions. Rather than relying on unguided exploration, it conditions on the critic’s current diagnoses to produce refinements that directly improve task reward.
    \vspace{-1mm}
    \item $C_\psi$ (The Diagnostic Evolver) is rewarded for feedback that maximizes the policy's performance gain, thereby learning to pinpoint the flaws that causes the policy's failure.
\end{itemize}
\vspace{-1mm}
This joint evolution ensures that the critic’s diagnostic depth is continuously calibrated to the policy’s shifting failure patterns. By optimizing both models through a dual-track GRPO mechanism, we transform the refinement process into a self-improving system where evaluative precision and execution capability evolve in tandem. 
Figure \ref{fig:method} summarizes the overall training loop and illustrates a concrete refinement example.

\subsection{Cascaded Evolutionary Rollout}

To facilitate the symbiotic optimization of both models, ECHO employs a cascaded rollout mechanism that generates group-structured trajectories through a diagnostic-and-corrective cycle.

\paragraph{Stage 1: Multi-view Diagnosis.} Given a query $q$, the policy $P_\theta$ first generates an initial trajectory $\tau_o \sim P_{\theta}(\cdot \mid q)$. To provide an objective basis for diagnosis, an external reward model $R$ evaluates the proposal to obtain a baseline score $s_o = R(q, \tau_o)$. Conditioned on both the trajectory and its corresponding score, the critic $C_{\psi}$ is invoked $N$ times independently to produce a set of diverse diagnostic feedbacks $\mathcal{G}_C = \{c_o^{(j)}\}_{j=1}^{N}$:
\begin{equation}
    c_o^{(j)} \sim C_{\psi}(\cdot \mid q, \tau_o, s_o), \quad j = 1, 2, \dots, N.
\end{equation}
By incorporating $s_o$ into the prompt, the critic is empowered to provide "score-aware" explanations, identifying the specific gaps that prevent the trajectory from achieving a higher reward.

\paragraph{Stage 2: Conditional Refinement.} Following the diagnosis, the policy $P_\theta$ is required to internalize these critiques into precise corrective actions. Conditioned on the augmented input $\tilde{q}^{(j)} = (q, c_o^{(j)})$, the policy samples a corresponding set of refined trajectories:
\begin{equation}
    \tau_r^{(j)} \sim P_{\theta}(\cdot \mid \tilde{q}^{(j)}), \quad j = 1, 2, \dots, N.
\end{equation}
The reward model evaluates each refinement to yield the post-correction scores $s_r^{(j)} = R(q, \tau_r^{(j)})$. This cascaded rollout produces the baseline score $s_o$, the critique group $\mathcal{G}_C$, and the refinement group $\mathcal{G}_P = \{\tau_r^{(j)}\}_{j=1}^N$, which serve as the empirical signals for the co-evolutionary optimization.

\subsection{Saturation-Aware Reward Design}\label{sec: reward_design}
A straightforward approach to quantifying the utility of a critique is to measure the linear improvement in reward, \textit{i.e.}, $\Delta s = s_r - s_o$. However, this linear metric fails to account for the \textit{saturation effect} in model optimization: as the initial score $s_o$ approaches the performance ceiling (\textit{e.g.}, $s \to 1$), the marginal effort and information required to achieve a further increment surge. 
Treating an improvement from $0.9$ to $0.95$ as equivalent to one from $0.1$ to $0.15$ creates an "equidistant fallacy," which discourages the critic from diagnosing subtle yet critical flaws in high-quality proposals and leads to optimization plateaus.

To address this, we hypothesize that the reward space is non-linear and governed by a difficulty weighting function $\omega(s)$. We define $\omega(s)$ as a soft barrier function that captures the increasing difficulty of entropy reduction as perfection is approached:
\begin{equation}
    \omega(s) = \frac{1}{1 - s + \eta},
\end{equation}
where $\eta > 0$ is a smoothing hyperparameter. We define the intrinsic gain of a refinement as the path integral of $\omega(s)$ from $s_o$ to $s_r$:
\begin{equation}
    g(s_o, s_r) = \int_{s_o}^{s_r} \omega(s) ds = \ln \left( \frac{1 - s_o + \eta}{1 - s_r + \eta} \right).
\end{equation}

This choice yields a principled shaping signal~\cite{ng1999policy} with three desirable properties.
First, it is \emph{saturation-aware}: for the same $\Delta s$, the gain $g$ is larger
when the improvement happens in a higher-score region, encouraging the critic to focus on subtle yet impactful flaws in near-correct proposals. Second, it is \emph{additive}
(path-consistent):
\begin{equation}
    g(s_o, s_m) + g(s_m, s_r) = g(s_o, s_r),
\end{equation}
which makes the training signal invariant to whether refinement is performed in one
step or through multiple intermediate edits. Third, the gain is \emph{antisymmetric},
$g(s_o,s_r) = -g(s_r,s_o)$, providing a unified measure that rewards
improvements and penalizes regressions under the same scale.

Finally, we use this intrinsic gain directly as the critic reward:
\begin{equation}
    \label{eq: reward}
    r_c = g(s_o, s_r)
    = \ln \left( \frac{1 - s_o + \eta}{1 - s_r + \eta} \right).
\end{equation}

\subsection{Synchronized Co-evolutionary Optimization}
Instead of treating the critic as a static oracle, we operationalize the co-evolution as a synchronized dual-track alignment problem. We formulate a closed-loop optimization where both $P_\theta$ and $C_\psi$ explore a shared trajectory space, mutually anchoring each other’s learning progress. This is achieved by constructing two interdependent group structures:
\begin{align}
    & \mathcal{G}_P(q) = \{\tau_r^{(1)}, \tau_r^{(2)}, \dots, \tau_r^{(N)}\}, \\
    & \mathcal{G}_C(q, \tau_o) = \{c_o^{(1)}, c_o^{(2)}, \dots, c_o^{(N)}\}.
\end{align}
Here, $\mathcal{G}_C$ represents the diagnostic hypothesis space containing $N$ distinct interpretations of the proposal’s flaws, while $\mathcal{G}_P$ represents the corrective action space conditioned on those hypotheses.

\paragraph{Dual-Track Advantage Estimation.} To maximize sample efficiency, we compute group-relative advantages that capture the marginal utility of each model’s output. For the policy $P_\theta$, the advantage $A_P^{(j)}$ is computed by normalizing the scores $s_r^{(j)}$ within $\mathcal{G}_P$. This allows the policy to efficiently identify the most effective refinement paths from diverse diagnostic samples~\cite{wang2022self,cobbe2021training}. For the critic $C_\psi$, the advantage $A_C^{(j)}$ is derived by performing group-relative normalization on the saturation-aware rewards $r_c^{(j)}$ defined in Section \ref{sec: reward_design}. By amplifying high-score gains and balancing penalties via $\lambda$, the mechanism enables the critic model to rapidly converge on effective feedback.

\paragraph{Synchronized Update.} Following the Group-Relative Policy Optimization (GRPO) objective~\cite{shao2024deepseekmath}, both $P_\theta$ and $C_\psi$ are updated by maximizing a surrogate objective that incorporates advantage-weighted likelihood and a KL divergence constraint:
{\scriptsize 
\begin{equation}
\label{grpo_loss}
\begin{aligned}
\mathcal{J}(\phi) = &\mathbb{E}_{q \sim \mathcal{D}, \{o_i\}_{i=1}^{N} \sim M_{\phi_{\text{old}}}} \Bigg[ \frac{1}{N} \sum_{i=1}^{N} \frac{1}{|o_i|} \sum_{t=1}^{|o_i|} \min \Bigl( \rho_{i,t}(\phi) A_{i}, \\
&\text{clip}(\rho_{i,t}(\phi), 1-\epsilon, 1+\epsilon) A_{i} \Bigr) \Bigg] - \beta D_{\text{KL}}(M_{\phi} \| M_{\text{ref}}),
\end{aligned}
\end{equation}
}
where $\phi \in \{\theta, \psi\}$ represents the parameters of the policy or critic, and $o_i$ denotes the generated sequence. The importance sampling ratio is defined as $\rho_{i,t}(\phi) = \frac{M_{\phi}(o_{i,t} \mid \text{ctx}, o_{i,<t})}{M_{\phi_{\text{old}}}(o_{i,t} \mid \text{ctx}, o_{i,<t})}$, where $\text{ctx}$ is the corresponding input context for each model. $A_i \in \{A_P, A_C\}$ is the respective group-relative advantage. This synchronized optimization ensures the critic’s diagnostic focus is continuously calibrated to the policy’s evolving failure patterns, fostering a self-reinforcing curriculum for continuous improvement. For completeness, the full pseudo-code of ECHO is provided in Appendix~\ref{appendix:code}.

\section{Experiment Setup}

\paragraph{Scenarios and tasks.} To evaluate ECHO across a broad spectrum of cognitive challenges, we conduct experiments in four diverse environments. Specifically, for web navigation, we use WebShop~\citep{yao2022webshop}, requiring agents to navigate e-commerce platforms and make purchasing decisions; for embodied tasks, ALFWorld \citep{shridhar2020alfworld} challenges agents with long-horizon planning and object manipulation in household settings; for scientific tasks, SciWorld \citep{wang2022scienceworld} provides a simulator for complex experimental reasoning and hypothesis verification; and for deep search, we adopt the RAG-based DeepSearch environment from \citet{xi2025agentgym}, which requires multi-turn information synthesis for open-domain question answering. More details are shown in Appendix \ref{appendix: tasks}.
\paragraph{Baselines and backbone models.} We utilize Qwen3-4B-Instruct-2507~\citep{yang2025qwen3} (denoted as Qwen3-4B in the following) and Qwen2.5-7B~\citep{team2024qwen2} as primary backbone models. By default, the critic $C_{\psi}$ uses the same backbone as the policy $P_{\theta}$.
To ensure a rigorous and comprehensive evaluation, we compare our method against a diverse set of strong baselines spanning both proprietary and open-source large language models. Specifically, for {proprietary models}, we include GPT series~\citep{achiam2023gpt}, Gemini-2.5-pro~\citep{comanici2025gemini}, and Claude-Sonnet-4.5. In addition, we consider Open-sourced Models as competitive baselines, including Qwen3-235B-A22B~\citep{yang2025qwen3} and DeepSeek-R1-0528~\citep{guo2025deepseek}. The implementation detail is described in Appendix~\ref{appendx: ex_detail}.

\section{Results}
Table~\ref{tab:main_results} presents the main results. 
We organize our analysis around three research questions: \textbf{RQ1} evaluates the overall effectiveness of ECHO on open-world agent benchmarks; \textbf{RQ2} investigates whether failure patterns drift during on-policy learning and whether this drift causes a frozen critic to become stale; and \textbf{RQ3} studies why the proposed saturation-aware reward is beneficial, especially for last-mile improvements near the reward ceiling. More detailed experimental results and analyses are provided in Appendix~\ref{sec:appendix_exp}.

\subsection{RQ1: How effective is ECHO for open-world agent learning?}

{\renewcommand{\arraystretch}{1.1}
{\setlength{\tabcolsep}{8pt}
\begin{table*}[!htbp]
\centering
\caption{Main results on four open-world agent benchmarks. \textbf{Bold} indicates the best result within each benchmark.}
\vspace{-1mm}
\label{tab:main_results}
\begin{tabular}{lccccc}
\toprule
\multicolumn{1}{c}{\textbf{Models}}             & {\textbf{WebShop}} & {\textbf{ALFWorld}} & {\textbf{SciWorld}} & {\textbf{DeepSearch}} & {\textbf{Overall}} \\ \hline
\rowcolor{gray!10}
\multicolumn{6}{c}{\textit{Proprietary Models}}                                  \\ 
GPT-4o-mini        & 56.59 & 45.20 & 40.68 &   31.43     &  43.48      \\
GPT-4o             & 58.20 & 44.45 & 45.78 &   26.19     &  43.66      \\
GPT-4-turbo        & 52.45 & 42.64 & 34.14 &   61.90     &  47.78      \\
GPT-4.1            & 58.07 & 43.56 & 35.65 &   61.46     &  49.67      \\
GPT-5              & 46.12   & 35.09    & 13.06    & 72.19         &  41.62       \\
Gemini-2.5-pro     & 65.58   & 68.04    & 12.50    & 36.50    & 45.66 \\
Claude-Sonnet-4.5 & 58.80   & 64.73    & 56.83    & 65.00          & 61.34       \\ \hline
\rowcolor{gray!10}
\multicolumn{6}{c}{\textit{Open-sourced Models $\ge 100B$}}                                 \\ 
Qwen3-235B-A22B    & 25.26   & 26.60    & 23.50    & 28.25    & 25.90    \\
DeepSeek-R1-0528   & 44.81   & 72.50    & 4.50     & 40.25    & 40.52    \\ \hline
\rowcolor{gray!10}
\multicolumn{6}{c}{\textit{Open-sourced Models < 100B \& RL}}                           \\ 
Qwen3-4B           & 6.12    & 0.32     & 4.50     & 20.25    & 7.80    \\
Qwen3-4B + GRPO    & 82.37   & 87.50     & 79.14     & 33.25      & 70.57        \\
Qwen3-4B + ECHO    & \textbf{90.03}   &  \textbf{91.25}     & \textbf{82.88}   & \textbf{47.25}  & \textbf{77.85}        \\ \hdashline
Qwen2.5-7B         & 13.98   & 2.00     & 1.50     & 15.50    & 8.25    \\
Qwen2.5-7B + GRPO  & 83.55  & 89.50     & 81.24      & 42.25      & 74.14       \\
Qwen2.5-7B + ECHO  & \textbf{89.97}  & \textbf{93.75}      & \textbf{85.63}      & \textbf{46.75}      & \textbf{79.03}       \\ 
\bottomrule
\end{tabular}
\end{table*}
\vspace{-1mm}

\paragraph{ECHO consistently outperforms standard GRPO and other strong baselines.}

As shown in Table~\ref{tab:main_results}, ECHO consistently surpasses GRPO under the same training budget, supporting our hypothesis that synchronized, on-policy critiques reduce unproductive exploration and thus improve data efficiency.
The most salient gains appear on Qwen3-4B in long-horizon search and web interaction: on {DeepSearch}, ECHO improves from 33.25 to 47.25, roughly a 42\% relative increase; on WebShop, it rises from 82.37 to 90.03, about a 9\% relative increase. These boosts indicate that ECHO is especially effective when success depends on diagnosing and repairing specific failure causes across multiple steps. 
Importantly, in more complex embodied and scientific environments where failures are more diverse and harder to localize, ECHO also brings consistent gains on Qwen3-4B, improving ALFWorld from 87.50 to 91.25 and SciWorld from 79.14 to 82.88.
Overall, ECHO improves performance across all four benchmarks, achieving an average gain of {7.28} points over GRPO, and it delivers highly competitive results against much stronger baselines: except for {DeepSearch}} where GPT-5 attains the best score, ECHO matches or surpasses all listed strong models by a clear margin on the other benchmarks.

\paragraph{ECHO generalizes across backbone sizes.}
To test whether ECHO is applicable across different backbone sizes, we also evaluate it on Qwen2.5-7B. The results show that ECHO is not restricted to a specific capacity regime. Instead, it consistently improves over GRPO on both backbones and yields strong performance across environments. This demonstrates that the benefit of synchronized critic-policy co-evolution transfers across model scales, highlighting the versatility and generalizability of ECHO for open-world agent learning.

\subsection{RQ2: Does fail-pattern drift happen during on-policy learning?}

{\renewcommand{\arraystretch}{1.0}
{\setlength{\tabcolsep}{6pt}
\begin{table*}[!htbp]
\centering
\caption{Ablation results of ECHO. ``w/o'' denotes removing the specified component. Since SA shaping relies on meaningful reward magnitudes, we focus on WebShop and SciWorld, two benchmarks with non-binary rewards. }
\vspace{-1mm}
\label{tab:ablation}
\resizebox{0.85\linewidth}{!}{%
\begin{tabular}{lccccc}
\toprule
\multicolumn{1}{c}{\textbf{Methods}}             & {\textbf{WebShop}} & {\textbf{ALFWorld}} & {\textbf{SciWorld}} & {\textbf{DeepSearch}} & {\textbf{Avg($\downarrow$)}} \\ \hline
\rowcolor{gray!10}
\multicolumn{6}{c}{\textit{Qwen3-4B}}                                  \\
GRPO                 & 82.37   & 87.50   & 79.14    & 33.25    & -    \\
ECHO                 & \textbf{90.03}   & \textbf{91.25}   & \textbf{82.88 }   & \textbf{47.25}    & -        \\
ECHO w/o evloving    & 83.60   & 85.75 & 68.58    & 40.25    & 9.25        \\
ECHO w/o SA-aware    & 86.69   & -   & 78.55    & -    & 3.84        \\
\rowcolor{gray!10}
\multicolumn{6}{c}{\textit{Qwen2.5-7B}}    \\
GRPO  & 83.55  & 89.50     & 81.24      & 42.25      &  -      \\
ECHO  & \textbf{89.97}  & \textbf{93.75}      & \textbf{85.63}      & \textbf{46.75}      &  -     \\ 
ECHO w/o evloving    & 84.99   & 92.50   & 72.19    & 42.50    & 5.98    \\
ECHO w/o SA-aware    & 86.78   & -  & 83.65    & -    & 2.59        \\

\bottomrule
\end{tabular}
}
\end{table*}
}}


\begin{figure*}[t]
  \centering
  \includegraphics[width=1\textwidth]{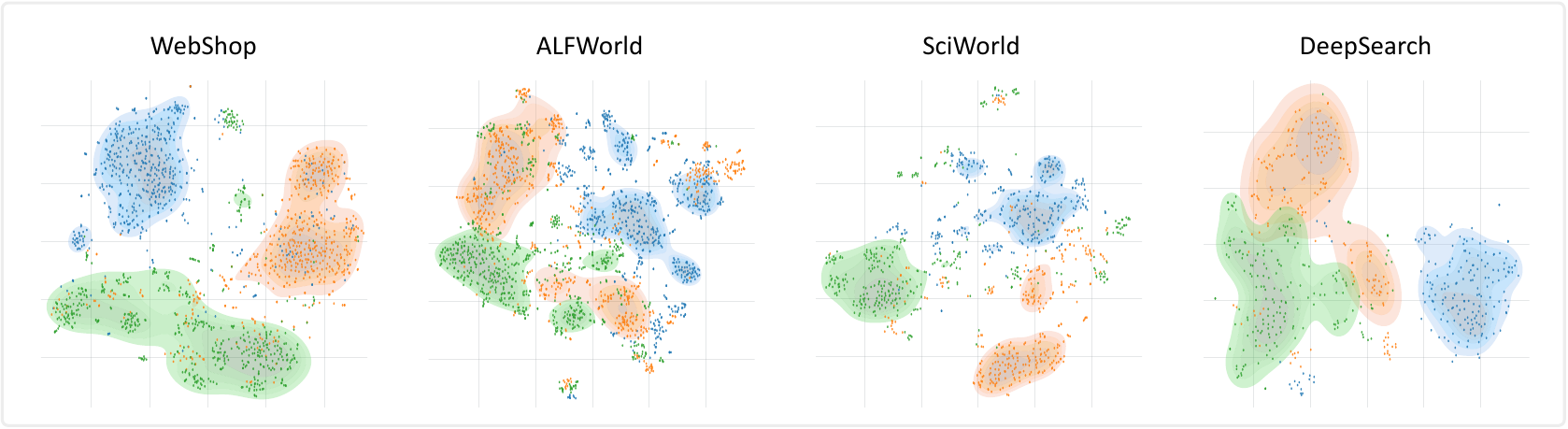}
  \vspace{-2mm}
  \caption{Failure-pattern drift across training phases. We visualize failed trajectories from \textcolor{myblue}{early}, \textcolor{myorange}{intermediate}, and \textcolor{mygreen}{late} checkpoints in a diagnosis embedding space using t-SNE, with contours indicating density regions.}
  \label{fig:dist_shift}
\end{figure*}

\subsubsection{How Failures Change Over Training}
To further examine whether failure patterns drift under on-policy training, we analyze the training trajectory of Qwen3-4B and partition it into three phases: early, intermediate, and late. 
In each phase, we select three adjacent policy checkpoints, and for every checkpoint we run rollouts on the same held-out test set. 
We collect all unsuccessful trajectories produced in each phase and treat them as samples from the phase-specific failure distribution.
For each unsuccessful trajectory, we use Gemini-2.5-pro to produce a concise diagnosis describing the underlying error cause. We then embed these diagnoses using Qwen3-8B-Embedding and visualize the resulting representations with t-SNE \citep{maaten2008visualizing}.

\paragraph{Phase-wise drift of dominant failure modes.}
Figure~\ref{fig:dist_shift} shows clear distributional drift across all four environments. 
In WebShop and DeepSearch, failures in each phase form relatively compact clusters, and the high-density centers shift substantially from early to late. This indicates that training changes which error causes dominate, rather than simply shrinking a fixed set of mistakes.

\paragraph{Higher diversity and partial persistence in complex environments.}
In the more complex environments ALFWorld and SciWorld, the failure distributions are more dispersed and partially overlap across phases, reflecting higher failure-mode diversity and the persistence of some recurring errors. 
Even in these settings, the density mass still migrates across training phases, confirming that the dominant failure patterns remain non-stationary.


\begin{figure*}[htbp]
  \centering
  \includegraphics[width=\textwidth]{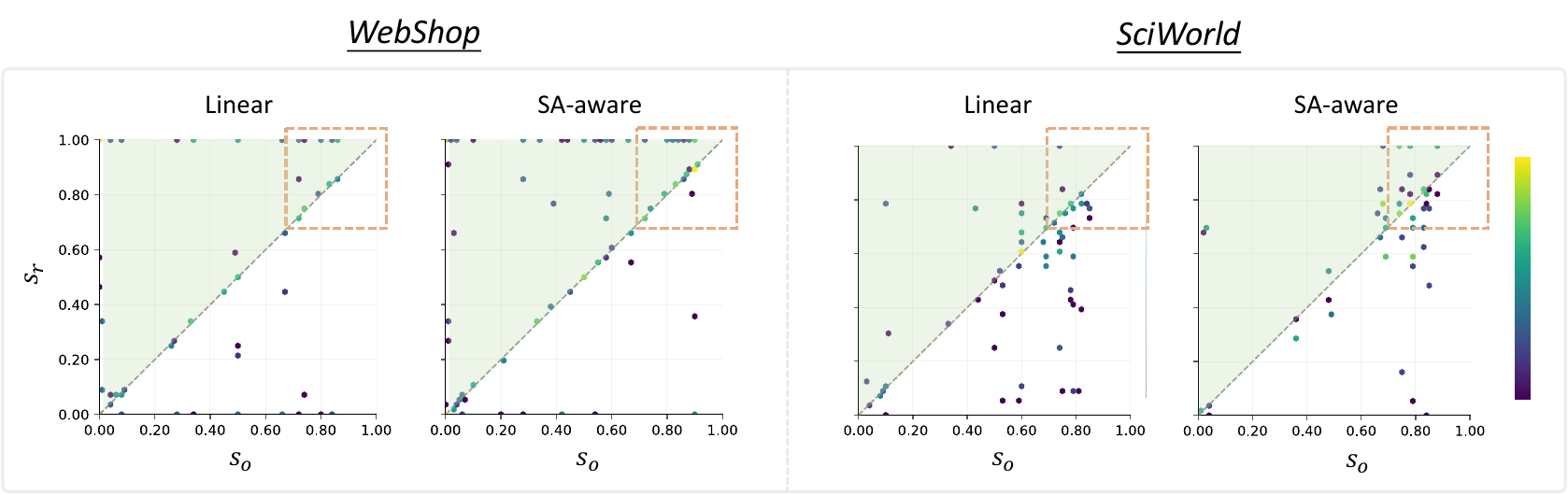}
  \vspace{-4mm}
  \caption{Effect of saturation-aware gain shaping on last-mile refinement. We plot density scatter maps of pre-refinement and post-refinement rewards \((s_o, s_r)\) on WebShop and SciWorld using Qwen3-4B. Points in the green region satisfy \(s_r > s_o\) and correspond to reward-improving refinements, where higher density indicates more effective critiques. The highlighted high-score square marks the near-ceiling regime.}
  \label{fig:reward}
\end{figure*}

\subsubsection{Limitations of Frozen Critics under Failure-Pattern Drift}




To further validate the need for critic adaptation under failure-pattern drift, we freeze the critic and rerun the experiments with all other components of ECHO unchanged. 
Results are presented in Table~\ref{tab:ablation} and illustrated by the training curves in Figure~\ref{fig:reward_curve}.

\begin{figure}[t]
  \centering
  \includegraphics[width=0.495\textwidth]{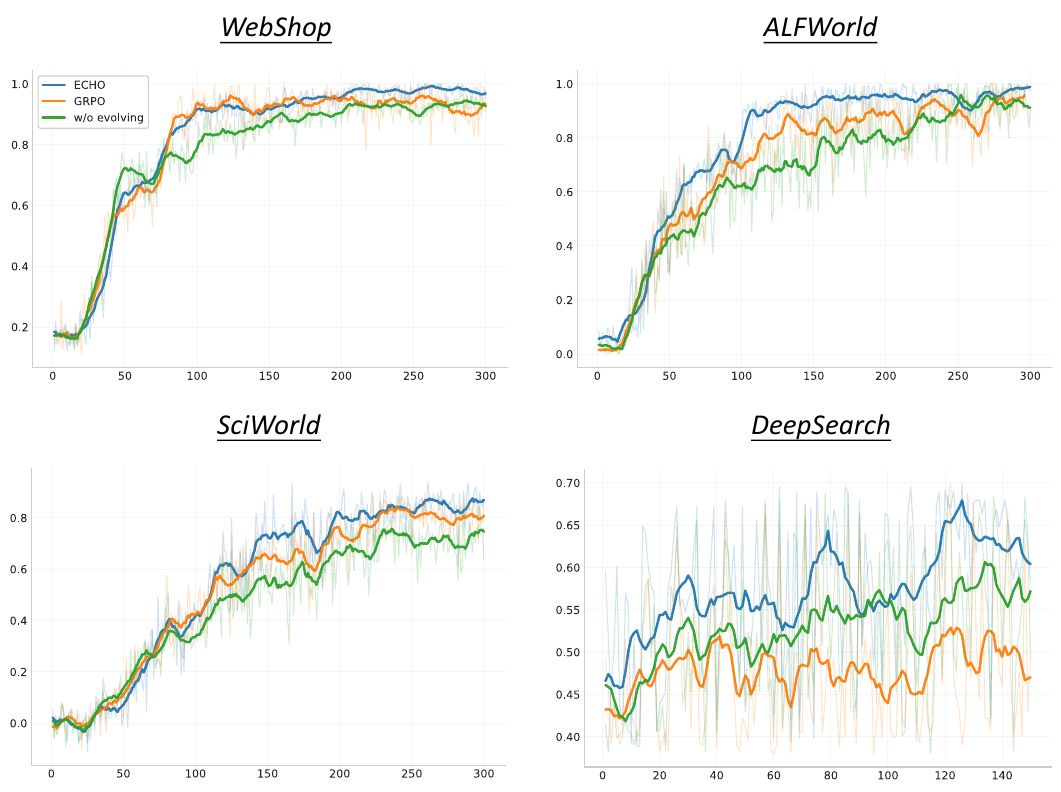}
  \vspace{-5mm}
  \caption{Training reward curves across four environments (Qwen3-4B).}
  \label{fig:reward_curve}
  
\end{figure}

\paragraph{Final performance drops with a frozen critic.}
 We find that this simple change leads to performance degradation across all environments, indicating that keeping critiques synchronized with the evolving policy is important for maintaining their effectiveness. 
 Meanwhile, the degradation is most severe on {ALFWorld} and SciWorld, and even underperform standard GRPO. We conjecture that in these more complex environments, a stale critic more frequently produces redundant or off-target diagnoses, which the policy may over-condition on during refinement, turning critiques into noise and amplifying long-horizon errors.

\paragraph{Training dynamics reveal phase-dependent effects.}
Figure~\ref{fig:reward_curve} further shows that the benefit of co-evolution depends on both training phase and environment.
On {WebShop}, the frozen-critic variant can look strong early on, but its improvement slows later and is overtaken by ECHO, consistent with later-stage errors becoming more fine-grained such that stale critiques are increasingly miscalibrated and act as noise that reduces sampling efficiency. 
In {ALFWorld} and {SciWorld}, ECHO stays close to GRPO at the beginning and separates mainly in the mid-to-late stage, suggesting a short calibration period in which the critic learns to produce environment-specific, actionable diagnoses for long-horizon failures before its advantage becomes visible.
By contrast, on {DeepSearch}, ECHO improves more steeply in the early stage; we hypothesize this is because the evaluator is highly sensitive to output format and interaction protocol, so the critic can quickly correct systematic, easy-to-specify early failures. 

Overall, these curves support our claim that critique strategies are non-stationary under on-policy training: as failure modes drift, a frozen critic becomes increasingly mismatched, whereas synchronized co-evolution helps maintain critique utility throughout training.


\subsection{RQ3: Why is the saturation-aware (SA) reward design effective?}

To examine whether SA gain shaping provides a more informative learning signal than a linear improvement reward, we compare two reward designs on Qwen3-4B in WebShop and SciWorld: the linear reward \(\Delta s = s_r - s_o\), and our saturation-aware gain \(g(s_o,s_r)\) in Eq.~\ref{eq: reward}. 

As shown in Table~\ref{tab:ablation}, disabling SA shaping while keeping the rest of ECHO unchanged leads to consistent drops on both datasets. Notably, the degradation is larger on WebShop. We attribute this to the different regimes reached by the policy: SciWorld is more challenging and the learned agent remains further from saturation, so training is less dominated by last-mile refinements where SA shaping is designed to provide extra signal; in contrast, WebShop more often enters a near-ceiling regime, making SA shaping more impactful.

To further understand {where} SA shaping helps during refinement, we next visualize the joint distribution of pre-refinement and post-refinement rewards \((s_o, s_r)\) in Figure~\ref{fig:reward}. Since saturation effects are most salient when trajectory rewards are already high, we focus on the middle-to-late stage of training. Specifically, we extract a window of 10 consecutive rollout batches, remove trajectories with \(s_o=1\), and visualize the joint distribution of \((s_o, s_r)\) as density scatter plots in Figure~\ref{fig:reward}.


\paragraph{Overall refinement effectiveness.}  
Across both WebShop and SciWorld, saturation-aware shaping concentrates substantially more probability mass in the improvement region where \(s_r > s_o\), shown as the green upper-left triangle in Figure~\ref{fig:reward}. Higher density in this region indicates that critiques more reliably translate into reward-increasing refinements, suggesting that the saturation-aware design yields stronger overall refinement effectiveness than the linear alternative.

\paragraph{Last-mile improvement near the reward ceiling.}  
In the high-score regime highlighted by the yellow square, the most desirable outcomes lie in its upper-left area, where trajectories start near full reward and still improve after refinement. For both datasets, saturation-aware shaping exhibits higher density in this region, indicating better ability to convert near-correct trajectories into full-reward solutions. In contrast, the linear reward shows many samples remaining close to the diagonal in this regime, especially on SciWorld, indicating that refinements tend to preserve the original score and struggle to achieve the small but critical gains required near the ceiling.




\section{Conclusion}
We presented ECHO, a co-evolution framework for open-world LLM agents. By synchronizing critic and policy updates, ECHO mitigates critic staleness under on-policy failure drift. The proposed cascaded rollout provides group-structured samples for group-relative optimization, while the saturation-aware gain shaping boosts last-mile improvements. Together, these designs enable the critic's diagnostic granularity to stay aligned with the policy's evolving failure modes, supporting more stable training and sustained refinement.

\section*{Limitations}
Our framework updates both the policy and the critic using improvement signals computed from the same external reward model. Therefore, its effectiveness depends on reward quality and calibration: if the reward is noisy, biased, or underspecified, the critic may optimize toward evaluator artifacts rather than truly diagnostic feedback, and the policy may inherit the same misalignment.

Moreover, reward evaluation and critique generation are handled by separate models in our current implementation. A natural next step is to unify them into a single model that both scores trajectories and produces actionable critiques, which could simplify the training pipeline and improve consistency between “what is rewarded” and “what is suggested.” We leave this integration to future work.

\bibliography{custom}

\appendix

\onecolumn

\section{Environments and Scoring Criteria}\label{appendix: tasks}
The evaluation environments used in our experiments are summarized in Table \ref{tab:env_overview}, including their task settings, the core abilities required of the agent, and the official scoring criteria.

\begin{table*}[!htbp]
\centering
\caption{Overview of evaluation environments. We summarize each environment's setting, the core abilities required from the agent, and the scoring criterion used by the official evaluator.}
\label{tab:env_overview}

\begin{tabular}{
  p{2.3cm}
  p{4.4cm}
  p{3.8cm}
  p{3.8cm}
}
\toprule
\textbf{Environments} & \textbf{Description} & \textbf{Required Agent Ability} & \textbf{Score Criterion} \\
\midrule
WebShop \citep{yao2022webshop} &
An interactive e-commerce website simulator where the agent navigates product pages and selects an item that satisfies a natural-language shopping goal. &
Goal parsing, web navigation, information retrieval, constraint tracking, reasoning over semi-structured fields. &
Purchase-based success, rewarding buys that match the requested product type and constraints (attributes/options/price). \\
\addlinespace

ALFWorld \citep{shridhar2020alfworld} &
A text-based embodied household environment derived from ALFRED, requiring multi-step manipulation (\textit{e.g.}, pick, place, clean, heat) to accomplish instructions. &
Subgoal decomposition, spatial and physical commonsense, hierarchical planning, multi-step action execution, instrument operation. &
Binary episode success judged by completion of the specified household task.\\
\addlinespace

SciWorld \citep{wang2022scienceworld} &
A simulated scientific discovery environment where the agent performs experiments, uses instruments, and reasons over observations to satisfy a scientific objective. &
Scientific skills, experimental design, instrument operation, causal and mechanistic reasoning. &
Progress-based scoring that rewards completing required main subgoals in order (with optional bonus goals), while any out-of-order main step triggers a failure score. \\
\addlinespace

DeepSearch \citep{xi2025agentgym} &
A retrieval-augmented multi-turn QA environment where the agent must search, read, and synthesize evidence to answer open-domain questions. &
Query decomposition, iterative retrieval, evidence aggregation, faithful synthesis, termination control. &
Answer-level correctness judged by the official QA evaluator (exact-match style).\\
\bottomrule
\end{tabular}
\end{table*}
\newpage

\section{More Implementation Details}
\label{appendx: ex_detail}
All experiments are conducted with sixteen H20-100GB GPUs. 
We use the same learning rate for both the policy $P_{\theta}$ and the critic $C_{\psi}$, setting
$\text{lr}_{\theta}=\text{lr}_{\psi}=1\times 10^{-6}$.
We set the rollout group size to $N=8$ by default, \textit{i.e.}, for each query we sample 8 independent
critiques from the critic and generate 8 corresponding refinements conditioned on these critiques. For the {policy} model, we follow the official setup for both reward design and evaluation protocols to ensure a fair and consistent comparison. For the {critic} model, we use the reward function in Eq.~(\ref{eq: reward}) and set the $\eta$ to 0.1 in all experiments. 

\section{Additional Experimental Analyses}\label{sec:appendix_exp}
We provides additional experimental analyses that complement the main results in the paper. 
Specifically, we first include comparisons with additional critique-guided baselines beyond the standard GRPO setup. 
We then analyze the effectiveness of critique-guided refinement, verify whether refinements genuinely follow the critic's diagnostic feedback, and study how critique granularity evolves during training. 
Finally, we report a detailed training time analysis to quantify the computational overhead introduced by ECHO. 
Unless otherwise specified, all experiments in this appendix are conducted using the Qwen3-4B model.

\subsection{Comparisons with Additional Critique-Guided Baselines}

To further strengthen the empirical evaluation of ECHO, we include additional comparisons with two representative critique-guided approaches beyond the standard GRPO baseline. Specifically, we consider {RCO}, a training-based critic optimization framework that iteratively refines trajectories using a learned critic, and {LUFFY}, a strong refinement method that leverages teacher-provided solutions during trajectory improvement. These methods represent two common paradigms of critique-guided learning: training-based and template-based critics.

Table~\ref{tab:baseline_compare} presents the performance comparison across four environments. ECHO consistently outperforms both baselines across all tasks. Notably, although LUFFY benefits from strong teacher hints derived from ground-truth solutions, ECHO still achieves higher performance without access to such privileged supervision. This result suggests that the improvements of ECHO arise not from stronger external guidance, but from the adaptive diagnostic feedback provided by the co-evolving critic, which remains aligned with the policy's evolving failure patterns.

We note that RCO performs comparatively worse in our setting primarily because it does not update the policy model itself, relying instead on an iterative refine-and-evaluate process with a fixed policy. Such a design is less effective in long-horizon interactive environments where policy adaptation is critical. Overall, these results further support our conclusion that critic co-evolution, rather than the choice of RL optimizer alone, plays a key role in driving the performance gains of ECHO.

\begin{table}[t]
\centering
\small
\caption{Performance comparison with established critique-guided baselines across four environments. }
\label{tab:baseline_compare}
\begin{tabular}{lccccc}
\toprule
\textbf{Method} & \textbf{WebShop} & \textbf{ALFWorld} & \textbf{SciWorld} & \textbf{DeepSearch} & \textbf{Overall} \\
\midrule
RCO   & 35.64 & 3.00  & 16.50 & 25.75 & 20.22 \\
LUFFY & 80.34 & 80.92 & 70.44 & 31.00 & 65.18 \\
\textbf{ECHO} & \textbf{90.03} & \textbf{91.25} & \textbf{82.88} & \textbf{47.25} & \textbf{77.85} \\
\bottomrule
\end{tabular}
\end{table}

\subsection{Effectiveness of Critique-Guided Refinement}

We first investigate whether the performance gains brought by ECHO truly arise from critic-guided refinement, rather than simply from additional trajectory sampling. To this end, we design a controlled comparison that isolates the effect of the critic's diagnostic feedback.

For each query, we first generate an initial trajectory using the current policy. We then produce a second-pass rollout under two conditions: (1) \textbf{Critic-Guided Refinement}, where the trajectory is refined based on the critic's diagnostic feedback, and (2) \textbf{No-Critic Regeneration}, where we simply re-sample trajectories without any critique. For both conditions, we generate $N=8$ trajectories and compute the group-average reward. We then subtract the reward of the original trajectory to obtain the \emph{reward gain} for that query.

Table~\ref{tab:refinement_gain} reports the average reward gains across early, intermediate, and late training checkpoints on WebShop and SciWorld. The \textit{No-Critic Regen} column measures the improvement obtained purely from additional sampling, providing a fine-grained estimate of the natural variability of trajectories. The \textit{Critic-Guided Refine} column reports the gains achieved when refinement is guided by the critic. Finally, \textit{Relative Gain} measures the additional improvement attributable to the critic beyond this baseline variance.

Across both datasets and all training phases, critic-guided refinement consistently yields substantially higher reward gains than no-critic regeneration, demonstrating that the critiques provide actionable guidance beyond simple trajectory re-sampling. Notably, the relative gain becomes larger in the late phase of training, suggesting that the critic becomes increasingly effective as the policy and critic co-evolve.

\begin{table}[t]
\centering
\small
\caption{Reward gain from second-pass refinement across training phases. 
Values represent the average reward gain relative to the initial rollout. 
Critic-guided refinement is compared with no-critic regeneration under the same decoding budget.}
\label{tab:refinement_gain}
\begin{tabular}{llccc}
\toprule
\textbf{Dataset} & \textbf{Phase} & \textbf{No-Critic Regen} & \textbf{Critic-Guided Refine} & \textbf{Relative Gain} \\
\midrule
\multirow{3}{*}{WebShop}
& Early & +0.81 & +6.54 & +5.73 \\
& Mid   & +0.59 & +5.42 & +4.83 \\
& Late  & +0.23 & +7.54 & +7.31 \\
\midrule
\multirow{3}{*}{SciWorld}
& Early & +0.32 & +2.82 & +2.50 \\
& Mid   & +0.43 & +3.65 & +3.22 \\
& Late  & +0.34 & +3.78 & +3.44 \\
\bottomrule
\end{tabular}
\end{table}

\subsection{Critique--Refinement Alignment and Granularity Evolution}

We next investigate whether the improvements from critique-guided refinement are causally attributable to the critic's diagnostic feedback, and further examine how the nature of such feedback evolves during training.

\paragraph{Critique--Refinement Alignment.}
To verify whether refinements genuinely address the issues identified in critiques, we employ a fixed external evaluator (Gemini-2.5-pro) to assess the alignment between critiques and refined trajectories. For each instance, the evaluator is provided with the original trajectory, the critique, and the refined trajectory. Rather than judging overall correctness, it performs a targeted audit to determine whether the refinement explicitly resolves the issue identified in the critique.

The evaluator assigns one of three labels: \textit{YES} (issue resolved), \textit{NO} (issue not addressed), or \textit{UNCLEAR}. We discard \textit{UNCLEAR} cases and report the percentage of \textit{YES} labels as the \emph{issue-addressed rate}. As shown in Table~\ref{tab:alignment_granularity}, the alignment improves substantially over training. In the early phase, the issue-addressed rate is around 75\%, indicating that the policy has not yet learned to reliably follow critiques. In contrast, in intermediate and late phases, the rate exceeds 90\% across datasets, suggesting that the policy increasingly learns to act on the critic's diagnostic feedback.

\paragraph{Evolution of Critique Granularity.}
While the above results establish that refinements increasingly follow critiques, they do not explain why critiques become more effective in later stages. To this end, we analyze how the granularity of critiques evolves during training.

We classify critiques into three categories, namely \textit{Coarse-level}, \textit{Mid-level}, and \textit{Fine-grained}, based on their primary intent using a fixed external evaluator. Coarse-level critiques provide general guidance, mid-level critiques focus on structured reasoning or subtask-level corrections, and fine-grained critiques identify precise errors or decision points. 

As shown in Table~\ref{tab:alignment_granularity}, a clear shift occurs across training phases: early-stage critiques are dominated by coarse-level guidance, while mid-level and fine-grained critiques become increasingly prevalent in later stages. In particular, fine-grained critiques grow substantially in the late phase, indicating that the critic increasingly focuses on precise error localization.

To better understand this transition, we manually inspect representative rollouts. In early training, the policy often struggles with basic environment interaction, and critiques primarily provide procedural guidance. As the policy improves, errors shift toward suboptimal planning and long-horizon reasoning, and critiques correspondingly evolve to target higher-level decision-making or subtle reasoning flaws.

Overall, these results suggest that improved alignment is not merely a byproduct of training, but is driven by a systematic shift in critique granularity. As failure modes become more fine-grained and abstract, the critic adapts to provide increasingly precise and actionable feedback, enabling sustained performance gains in later stages of training.

\begin{table}[t]
\centering
\small
\caption{Critique--refinement alignment and critique granularity across training phases (\%). }
\label{tab:alignment_granularity}
\begin{tabular}{llcccc}
\toprule
\textbf{Dataset} & \textbf{Phase} & \textbf{Issue Addressed} & \textbf{Coarse-level} & \textbf{Mid-level} & \textbf{Fine-grained} \\
\midrule
\multicolumn{6}{c}{\textit{Critique--Refinement Alignment}} \\
\midrule
\multirow{3}{*}{WebShop}
& Early        & 74.56 & 62.42 & 24.12 & 13.46 \\
& Intermediate & 93.87 & 34.70 & 41.89 & 23.41 \\
& Late         & 95.30 & 8.61  & 49.34 & 42.05 \\
\midrule
\multirow{3}{*}{SciWorld}
& Early        & 75.15 & 68.90 & 20.66 & 10.44 \\
& Intermediate & 92.48 & 39.23 & 37.58 & 23.19 \\
& Late         & 90.02 & 11.78 & 41.50 & 46.72 \\
\bottomrule
\end{tabular}
\end{table}

\subsection{Training Time Analysis}

We analyze the computational overhead of ECHO to assess whether the performance gains come at a significantly increased training cost. In particular, we provide a detailed breakdown of wall-clock training time to identify the primary sources of overhead.

Table~\ref{tab:time_analysis} reports the training time decomposition on Qwen3-4B across multiple environments. For ECHO, we break down the total time into three components: policy rollout, critic rollout, and refinement. For the baseline GRPO, we report the rollout time and total training time under the same experimental setup.

The overhead introduced by ECHO's additional stages, which include initial trajectory generation, critic evaluation, and updates, is marginal. Compared to the total training time of the baseline GRPO, these costs are negligible, proving that the co-evolution mechanism itself isn't a bottleneck.

Instead, the bulk of the extra computation comes from the refinement stage. This is a logical trade-off: because refinement processes longer contexts, decoding naturally takes more time. Even with this refinement step, the impact on total wall-clock time remains manageable. We observed an average increase of roughly 15\% over GRPO, representing a modest trade-off given the substantial performance gains reported in our main experiments.
Note that in more demanding environments like ALFWorld and DeepSearch, training times do climb for both ECHO and GRPO. This is driven by the complexity of the tasks, not by the co-evolution logic itself.

Ultimately, these results show that ECHO's success isn't just the product of a massive compute budget. By delivering significantly stronger performance for a well-defined, moderate increase in time, ECHO offers a highly favorable efficiency trade-off.

\begin{table}[t]
\centering
\small
\caption{Training wall-clock time breakdown (seconds). 
We report the decomposition of training time for ECHO and compare it with the baseline GRPO under the same setup.}
\label{tab:time_analysis}
\begin{tabular}{lcccc}
\toprule
\textbf{Component} & \textbf{WebShop} & \textbf{ALFWorld} & \textbf{SciWorld} & \textbf{DeepSearch} \\
\midrule
Policy Rollout (ECHO) & 10  & 14  & 10  & 17  \\
Critic Rollout     & 10  & 8   & 9   & 10  \\
Refinement            & 272 & 541 & 192 & 581 \\
\textbf{Total ECHO}   & 327 & 592 & 231 & 649 \\
\hdashline
GRPO Rollout          & 259 & 502 & 168 & 552 \\
\textbf{Total GRPO}   & 273 & 519 & 184 & 581 \\
\hdashline
\textbf{Overhead (\%)} & +19 & +14 & +25 & +11 \\
\bottomrule
\end{tabular}
\end{table}

\section{Pseudo-code for ECHO}

Algorithm~\ref{alg:echo} provides the complete training procedure of ECHO. It summarizes the cascaded rollout pipeline (on-policy proposal $\tau_o$, multi-view critiques $\{c_o^{(j)}\}$, and critique-conditioned refinements $\{\tau_r^{(j)}\}$), the saturation-aware critic reward computation in Eq.~(\ref{eq: reward}), and the synchronized dual-track GRPO updates for the policy and the critic performed on the same on-policy batch.

\label{appendix:code}

\begin{algorithm}[H]
\caption{ECHO: Evolving Critic for Hindsight-Guided Optimization}
\label{alg:echo}
\DontPrintSemicolon
\SetKwInOut{Input}{Input}
\SetKwInOut{Output}{Output}

\Input{Dataset $\mathcal{D}$; reward model $R$; policy $P_\theta$; critic $C_\psi$; group size $N$; GRPO hyperparams $(\epsilon,\beta)$; smoothing $\eta>0$.}
\Output{Updated parameters $(\theta,\psi)$.}

\ForEach{training step}{
  Sample a batch of queries $q \sim \mathcal{D}$\;

  \BlankLine
  \tcp{Stage 0: On-policy proposal and baseline score}
  Sample initial trajectory $\tau_o \sim P_\theta(\cdot \mid q)$\;
  Compute baseline score $s_o \leftarrow R(q,\tau_o)$\;

  \BlankLine
  \tcp{Stage 1: Multi-view diagnosis (critic group)}
  \For{$j \leftarrow 1$ \KwTo $N$}{
    Sample critique $c_o^{(j)} \sim C_\psi(\cdot \mid q,\tau_o,s_o)$\;
  }
  $\mathcal{G}_C \leftarrow \{c_o^{(j)}\}_{j=1}^{N}$\;

  \BlankLine
  \tcp{Stage 2: Conditional refinement (policy group)}
  \For{$j \leftarrow 1$ \KwTo $N$}{
    Form augmented input $\tilde{q}^{(j)} \leftarrow (q, c_o^{(j)})$\;
    Sample refinement $\tau_r^{(j)} \sim P_\theta(\cdot \mid \tilde{q}^{(j)})$\;
    Evaluate post-correction score $s_r^{(j)} \leftarrow R(q,\tau_r^{(j)})$\;
  }
  $\mathcal{G}_P(q) \leftarrow \{\tau_r^{(j)}\}_{j=1}^{N}$\;

  \BlankLine
  \tcp{Saturation-aware critic reward for each critique}
  \For{$j \leftarrow 1$ \KwTo $N$}{
    $r_c^{(j)} \leftarrow \ln\!\Big(\frac{1-s_o+\eta}{1-s_r^{(j)}+\eta}\Big)$ \tcp*{Eq.~\ref{eq: reward}}
  }

  \BlankLine
  \tcp{Dual-track group-relative advantage estimation}
  Compute policy advantages $\{A_P^{(j)}\}_{j=1}^{N}$ by group-relative normalization of $\{s_r^{(j)}\}_{j=1}^{N}$\;
  Compute critic advantages $\{A_C^{(j)}\}_{j=1}^{N}$ by group-relative normalization of $\{r_c^{(j)}\}_{j=1}^{N}$\;

  \BlankLine
  \tcp{Synchronized GRPO updates (same batch, two tracks)}
  Update $\theta$ by maximizing the GRPO surrogate objective $\mathcal{J}(\theta)$ using sequences $\{\tau_r^{(j)}\}$ with advantages $\{A_P^{(j)}\}$\;
  Update $\psi$ by maximizing the GRPO surrogate objective $\mathcal{J}(\psi)$ using sequences $\{c_o^{(j)}\}$ with advantages $\{A_C^{(j)}\}$\;
}
\end{algorithm}

\section{Prompt for Critic Model}
\label{appendix: prompt}

We provide the exact prompting template used to elicit critiques from the critic model in Box~\ref{prompt_for_critic}. The prompt constrains the critic to ground its feedback in the official scoring information and to output at most 1--2 high-level, actionable suggestions in a fixed format, which stabilizes training and keeps critiques consistent across rollouts.

\begin{tcolorbox}[breakable, title = {Box A.1 \refstepcounter{mybox}\label{prompt_for_critic}: The prompt for critique generation.}]
You are a **Critic model**, used to provide guidance on a model's overall performance on a given task. \\

The system will provide you with: \\
- The user's task description; \\
- Several rounds of the model's interactions with the environment as it attempts to complete the task; \\
- The official final scoring. \\

Your task: **Strictly based on the official scoring information**, output clear improvement suggestions and behavioral guidance inside the <critic> tag. Only point out issues and directions for improvement. Do not mention strengths, and do not give praise or encouragement. \\

\texttt{\#\#} Input Description \\
Each round of the model's actions is provided within `<model\_response>...</model\_response>`. \\
Environment feedback is provided within `<env\_feedback>...</env\_feedback>`. \\

\texttt{\#\#} Official Scoring Criteria (for your understanding; do not modify or question) \\

\{Detail scoring criteria on specific task.\} \\

You must: \\
- Treat the official scoring result as absolutely correct and final.   \\
- Not reinterpret, question, or adjust the official score.  \\

\texttt{\#\#} Original Question \\
\{original\_prompt\} \\

\texttt{\#\#} Response to be Evaluated \\
\{initial\_response\} \\

\texttt{\#\#} Official Scoring Information \\
\{score\_info\_text\} \\

\texttt{\#\#} Output Format \\
You must output **only** the following two tags and their contents. Do not add or remove tags, and do not output anything outside these tags: \\
\textasciigrave\textasciigrave\textasciigrave \\
<reason> \\
Your detailed reasoning process: \\
  - Based on the official scoring information, analyze the model's performance on each scoring dimension; \\
  - Strictly follow the official scoring information; do not question, revise, or supplement it; \\
  - Whenever you have any doubts or subjective judgments about any evaluation dimension, always take the official scoring conclusion as the final basis. \\
</reason> \\
<critic> \\
Your final guidance: \\
  - If the model's score is a full score (1 point), directly return "none". \\
  - Give **at most 1–2** brief, high‑level suggestions; only mention the most critical issues. \\
  - Do **not** refer to specific reply text or dialogue details; only describe how model should change its general behavior or strategy. \\
  - Do not give any praise or encouragement; only point out problems and how You should improve. \\
  - Always address the model as **"You"** (second person), not "the model", "it", or similar. \\
</critic> \\
\textasciigrave\textasciigrave\textasciigrave
Based on the above standards, provide guidance on the given model's behavior, and output in the specified format. \\

\texttt{\#\#} Output \\
Now, begin your reasoning!
\end{tcolorbox}

\end{document}